# Tactile Sensing with a Tendon-Driven Soft Robotic Finger*

Chang Cheng
School of Biological Sci. and Medical Engr.
Beihang University
Beijing, China
Dept. of Math. and Computer Sci.
Colorado College
Colorado, USA
dcheng728@gmail.com

Yadong Yan
School of Biological Sci. and Medical Engr.
Beihang University
Beijing, China
adam7217@qq.com

Mingjun Guan
School of Biological Sci. and Medical Engr.
Beihang University
Beijing, China
mingjunguan7@163.com

Jianan Zhang
School of Biological Sci. and Medical Engr.
Beihang University
Beijing, China
baby0303zjn@buaa.edu.cn

Yu Wang
School of Biological Sci. and Medical Engr.
Beihang University
Beijing, China
wangyu@buaa.edu.cn

*Abstract*—In this paper, a novel tactile sensing mechanism for soft robotic fingers is proposed. Inspired by the proprioception mechanism found in mammals, the proposed approach infers tactile information from a strain sensor attached on the finger's tendon. We perform experiments to test the tactile sensing capabilities of the proposed structures, and our results indicate this method is capable of palpating texture and stiffness in both abduction and flexion contact. Under systematic cross validation, the proposed system achieved 100% and 99.7% accuracy in texture and stiffness discrimination respectively, which validate the viability of this approach. Furthermore, we used statistics tools to determine the significance of various features extracted for classification.

## I. Introduction

Soft materials have recently been incorporated in bionic hands more and more frequently for various advantages over rigid hands [1] [2] [3] [4]. One of the challenges for existing soft robotic hands is retrieving adequate sensory information. For a human, neuronal receptors are distributed all over their hand at different depth. These receptors are sensitive to tactile signals of different frequency; together, they help humans infer contact force, object stiffness, object texture, and other properties of the manipulation tasks [5]. Therefore, soft robotic hands require adequate tactile sensing capabilities to mimic the sensory-motor feedback system found in humans.

Past studies have employed rigid fingers to measure contact stiffness [6] [7]; these approaches estimate contact force and displacement based on the finger's kinematic model to calculate contact stiffness. However, his procedure is difficult to be transferred to a soft robotic finger: due to the compliant nature of soft manipulators, kinematic modeling is difficult and often less accurate. Existing work in soft manipulator tactile feedback can be classified by how the sensors are installed: through surface sensors and through embedded sensors [8]. The mechanism behind surface sensors is mimicking behaviors of surface mechanoreceptors, skin cells responsible for cutaneous perception [9]. Thus, studies taking this approach tend to combine neuromorphic signal processing methods with machine learning models [10] [11] [12]. For example, Rongala, Mazzoni, and Oddo fed outputs of a biomimetic tactile sensor to a neuronal activation heuristic function, and classified ten materials with accuracy of 97%. Human fingers contain receptors beyond the surface as well, thus an alternative approach to texture sensing is to utilize embedded sensors. This approach was explored by Hosada as they embedded pressure-sensitive sensors under a rubber fingertip, and demonstrated that the fingertip can learn to detect slippage [14]. Recent work by Zhao et al. explored a novel approach to achieving tactile feedback on a soft robotic hand [15]. They installed optically innervated waveguides in a pneumatically actuated finger; as the finger bends, the waveguides deform which results in power loss of the light within. Through analyzing the power loss, they found linear relationship between the top and bottom waveguide power loss when palpating objects with varying stiffness.

Aside from mechanoreceptors that respond to outside cues, human fingers retain proprioception, awareness of joint position. Neuronal agents of proprioception usually reside within the muscles, tendons, and joints [16]. The muscle and joint proprioceptive receptors provide positional feedback while the tendon proprioceptors are sensitive to strain. Therefore, humans' proprioception framework provide an alternative possibility to tactile sensing. This was explored by Homberg, Katzschmann, Dogar and Rus as they embedded flexible bending sensors in a pneumatically actuated soft robotic gripper [17]. With internal sensing capabilities, this gripper is capable of statically estimating its posture and identifying grasped objects.

*Research supported by "National Key R&D Program of China" under Grant 2017YFA0701101.

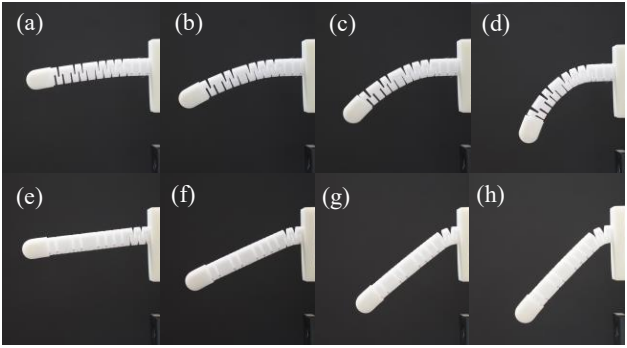

Fig. 1. The soft robotic finger under different strains. In (a-d), the finger is performing flexion under strains of 0N, 5N, 10N, and 15N, respectively. In (e-h), the finger is performing abduction under strains of 0N, 1N, 2N, and 3N, respectively. Note that the finger deforms slightly when no force is applied due to gravity (a, e).

In this paper, we aim to explore the possibility of texture and stiffness recognition by gathering temporal tactile feedbacks using the proprioception framework. Building on our previous work, we propose a biomimetic tendon-driven soft finger with a strain sensor attached. This design is combined with frequency domain analysis and machine learning models to perform texture and stiffness discrimination between eight and five categories, respectively. We then objectively evaluate the tactile sensing capabilities of the proposed system at different contacts. The resultant classification accuracies of 100%, 100%, 99.4%, and 99.7% at four contact modes indicate feasibility of the proposed system and that it does not require precise contact. Furthermore, we employ statistics tools to identify what features are key to the classification.

## II. DESIGN AND FABRICATION

### A. Soft Finger

The soft finger in this work is adopted from the TN hand [18]. It is a continuum structure with notches to allow deformation when actuated. The continuum's peculiar notch layout [19] allows flexion and extension of the finger, while little elongation or compression along the axial direction is permitted. The finger is fabricated with nylon using selective laser sintering technologies. The steel tendon that drives the finger has diameter of 0.5 mm. A cap fabricated with polyurethane is attached on the finger's end to mimic the human fingertip. Fig. 1 displays flexion and abduction motions of the finger under different tendon strains.

### B. Palpation Unit

The palpation unit consists of the finger, an actuator, a passive linear guide, an ATI 6-axis force/torque sensor, and the steel tendon. Ordinarily, the tendon passes through the finger and is directly connected to the actuator. Here, an ATI force/torque sensor is installed halfway to measure strain on the tendon. Only force in the axial direction of the 6-axis force/torque sensor is recorded, thus it functions as an 1-axis strain sensor. Directly below the sensor, a low-friction passive linear guide is installed to minimize gravity's influence on the sensor. The palpation unit is installed onto an active linear guide driven by a stepper motor (Fig. 2), which will slide the finger to palpate the textures.

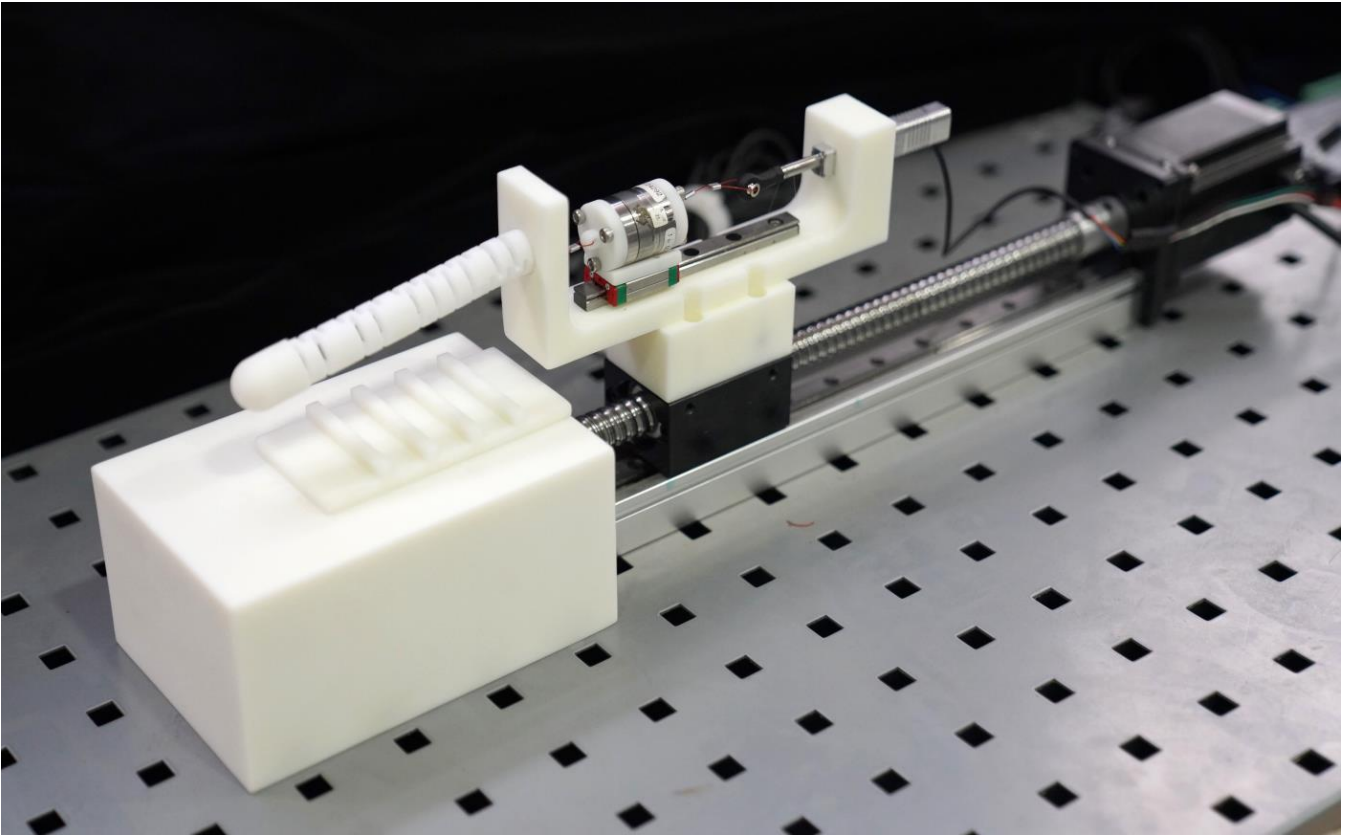

Fig. 2. The testing platform consisting of the palpated texture plate (left), palpation unit (center), and an active linear guide driven by stepper motor (right).

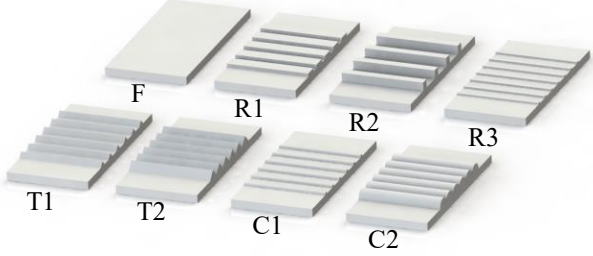

Fig. 3. The palpated texture plates.

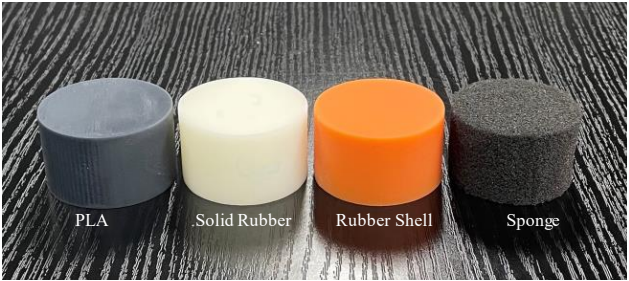

Fig. 4. The varied-stiffness cylinders for trial. An additional stiffness level of no contact is performed in the experiment.

*C. Tested Objects*

Eight texture plates and four varying-stiffness cylinders are trialed. The plates include flat surface (F), rectangular grooves (R1-R3), triangular grooves (T1, T2), and circular grooves (C1, C2). The plates are displayed in Fig. 3. Five levels of stiffness are trialed: PLA, solid rubber, rubber shell, sponge and no contact. These objects are displayed in Fig. 4.

## III. METHODS

The proposed framework's tactile sensing capabilities at different contacts are explored. A testing platform consisting of the palpation unit, an active linear guide, and palpated texture is built for this purpose and shown in Fig. 2. We divided the texture and stiffness sensing trials into groups of flexion contact (FC) and abduction contact (AC). In the FC trial groups, the texture plates are placed perpendicular to the finger's plane of flexion, and contact with the plate is created at the center of the fingertip as the finger bends. In the AC groups, the fingers are installed in abduction mode (Fig. 1 (e-h)) on the palpation unit, and the finger abducts to create contact on the fingertip's side.

*A. Data Acquisition*

As the trial begins, the active linear guide sends the palpation unit to where the fingertip is directly above the plate (Fig. 5 (a)). The finger bends roughly 40 degrees to create contact with the texture plate (Fig. 5 (b)). The active linear guide then glides the fingertip across the textured region at the rate of 15mm/second, and the tendon strain is registered at 60Hz (Fig. 5 (c-d)). The trial is complete after reaching the end of the plate. A total of 60 trials are executed on each texture plate for both FC and AC groups. For the stiffness trial groups, a tested object is first placed directly below the finger while tendon is relaxed (Fig. 5 (e)). Then the finger is actuated to apply pressure to the object. This state is kept for four seconds before the tendon is released and the finger returns to relaxed state (Fig. 5 (f-g)).

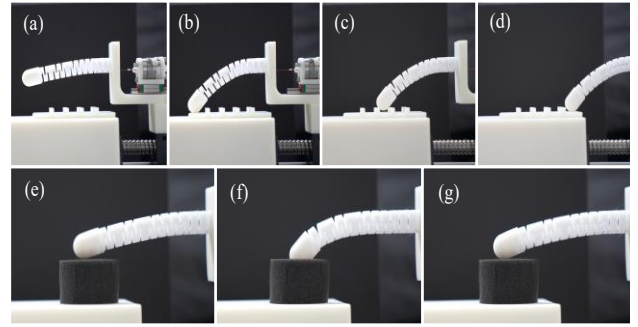

Fig. 5. From left to right, executions of the experiment for both texture (a-d) and stiffness (e-g) groups are displayed in sequence. While the finger is performing flexion in the above diagrams, the same executions are carried out for abduction groups.

*B. Feature Extraction*

Before extracting texture features, some preprocessing is needed. The beginnings and ends of each trial are cropped out as they represent uniformly flat sections. The next 180 continuous samples are taken, which amount to three seconds of gliding on the grooves. Lastly, zero-mean normalization is performed on the data to compensate the initial contact strain on the tendon. The resulting data would indicate the normalized fluctuation of the tendon stress when the finger palpates different plates. We then extract Fourier components of the post-preprocessing data. The magnitude of the Fourier components between the frequencies of 0 and 30Hz, with an interval of 0.33 Hz are used as features for classification. A sample of the raw strain data and extracted Fourier components are displayed in Fig. 6.

The stiffness data only contain strain progression from a single tap; it is not sequential thus requires a different feature extraction procedure. When the stiffness trials are performed, three phases occur: dynamic deformation, static deformation, and strain release. Dynamic deformation occurs as the finger initially reaches the palpated object. The palpated object undergoes a rapid deformation due to force applied by the finger and the strain increases dramatically. Then, as the finger holds the bent position for the next four seconds, it is still applying pressure to the object, which causes static deformation on the object, and the tendon's strain gradually decreases. The final strain release phase refers to when the finger leaves the object, and the tendon returns to a relaxed state. Because the dynamic deformation and strain release stages occur in a briefly time window, we determined extracting features from the static deformation phase shall result in more consistent data. Therefore, we perform linear regression to the static deformation phase of tendon strain with respect to time, and the resultant slope, intercept, and correlation coefficients are extracted as features. Respectively, these features indicate the decreasing rate, initial state, and linearity of the strain in the static deformation phase. Fig. 7 displays strain data collected during one stiffness trial.

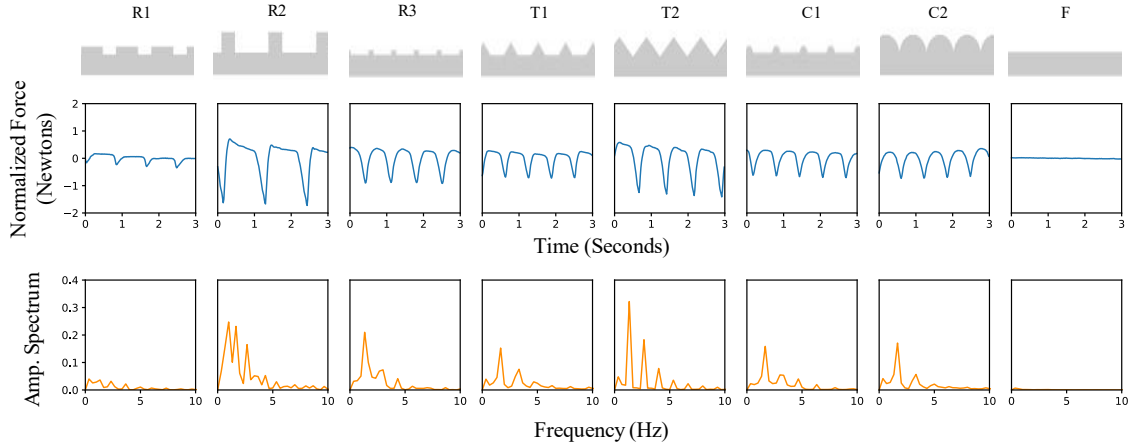

Fig. 6. Features for various textures in both time domain (middle) and frequency domain (bottom). The time domain forces have been normalized.

### C. Classification Algorithms

Four classification algorithms are used in this study: K Nearest Neighbor (KNN), Support Vector Machine (SVM) with linear kernel, SVM with radial basis function (RBF) kernel, and Decision Tree (DT). The KNN model implements Euclidean distance, and three nearest neighbors are voting.

## IV. RESULTS AND DISCUSSION

Fig. 6 displays the normalized strain and Fourier decompositions for trialed textures. The force data between R3, T1, T2, C1, C2 are quite difficult to distinguish by eye, but their difference become more apparent when transformed into frequency domain. The tendon strain during one stiffness trial is shown in Fig. 7, and the periods of dynamic deformation, static deformation, and strain release phases have been tagged.

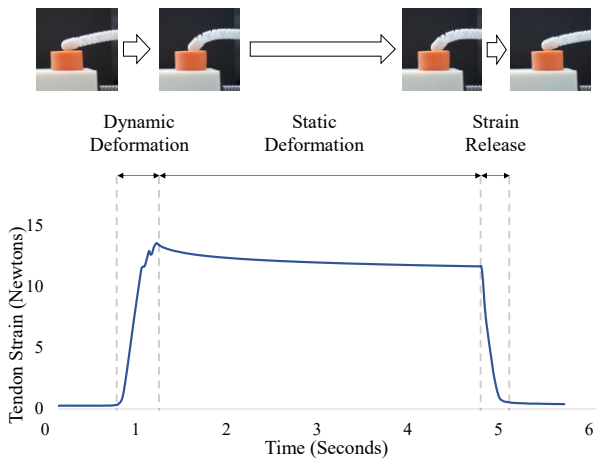

Fig. 7. The tendon strain during a FC stiffness trial. The corresponding regions of dynamic deformation, static deformation and strain release are displayed, and images above the graph indicates transitions between these phases. The slope, intercept, and correlation coefficient extracted from this trial are: -0.90, 12.62, -0.99.

To validate the classification results systematically, we utilized the $k$-fold cross validation. This method first pseudo-randomly partitions the dataset into $k$ subsets, or folds. Let there be folds of $f_1, f_2, ..., f_k$. As the validation begins, the learning model first omits $f_1$ and trains on $f_2$ to $f_k$, it then validates on the $f_1$ set and receive an accuracy. Afterward, the learning model resets and takes out $f_2$ as validation set, train, then validate again. This process is repeated for each of the folds, and the final accuracy is calculated by averaging the accuracy at each fold. The $k$-fold cross validation is often used to avoid bias or intentional training and validation set selection for embellishing machine learning performance. In our experiment, $k$ is set at six. We run the $k$-fold cross validation ten times and average the results to get an overall accuracy.

The classification accuracies of the proposed system using various methods is presented in Fig. 8. For the texture groups, the SVM with RBF kernel and KNN models consistently achieves 100% accuracy while the DT stably reaches high accuracy as well. In the stiffness groups, the linear SVM, KNN, and DT can reach above 97.2% accuracy while the SVM with RBF's accuracy is dramatically lower. The KNN and DT classifiers consistently achieve high accuracies in texture and stiffness discrimination for both contacts while the performance of the SVM classifier seem to depend on its kernel. This may imply that the stiffness dataset is linearly separable while the texture dataset is not. Equally high classification accuracies are achieved in both AC and FC results across various classifiers, which shows that the proposed system does not require precise contacts for texture and stiffness discrimination.

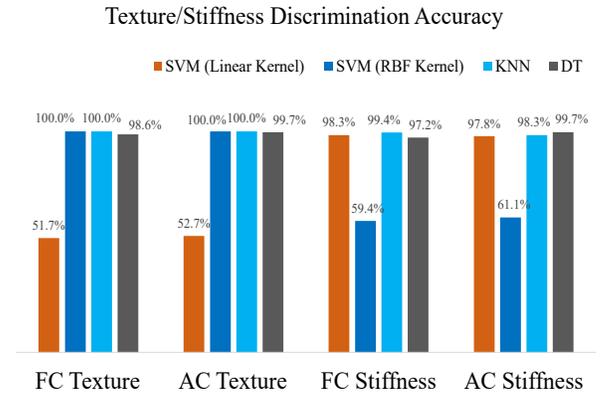

Fig. 8. The accuracies calculated for different contacts under four machine learning models. The accuracy at each column is the average of ten runs of k-fold cross validation.

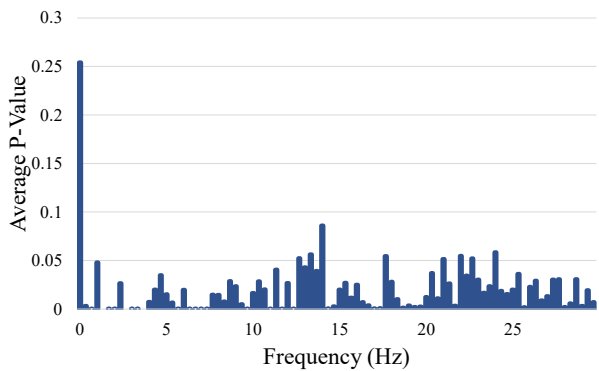

Fig. 9. Average p-value calculated for various frequencies. A low value in this chart would imply a high contribution to the classification. Besides features at frequency of 0Hz, only few Fourier components have average p-value above 0.05.

The input features for the classification algorithms are the magnitudes of decomposed Fourier components at different frequency. To understand what features in the frequency domain contributed more to the classification, we defined a significance measure for features. Let $p_i$ be the p-value matrix calculated by taking out the $i$-th Fourier component of each class, and running the Wilcoxon Rank Sum test between every combination of them. The classification significance measure of the $i$-th Fourier component is then calculated by averaging elements of $p_i$. Since the Rank Sum function determines the probability that two data series come from the same distribution, the p-value of the $i$-th Fourier component between two classes would imply the likelihood to separate these two classes by thresholding at a single value. Therefore, the average of the p-value matrix is an indicator for heterogeneity within a feature, and provides an estimate of the feature's contribution to classification. The results of the features' classification significance analysis is shown in Fig. 9. For future studies, selecting Fourier components that have relatively low average p-values in Fig. 9 as features may yield even better classification results. For the stiffness features, the p-value matrices are presented in Fig. 10, which shows that all three features selected are fairly dissimilar across the classes, except for the correlation coefficients between rubber and PLA.

## V. CONCLUSION

In this paper, a novel soft finger tactile sensing system is proposed, and its viability is experimentally validated with texture and stiffness discrimination accuracies between 97.2% and 100%. Unlike traditional methods toward finger tactile sensing, our design embeds the sensor in the tendon, which mimics the mammalian proprioceptive framework. This has the advantage of not requiring precise contacts for tactile feedback, which is backed by equally high discrimination accuracies between abduction and flexion contacts (Fig. 8). Our work is different from [17] in that [17] embedded bend sensors to mimic the joint proprioceptors that provide positional feedback, we mimicked the tendon proprioceptors which provide strain feedback. We also aimed at achieving texture and stiffness recognition while [17] investigated object recognition.

In addition to texture and stiffness classification, slippage and rolling detection has been marked as a challenge for robotic hands [21], which may be a topic to explore in the future using the proposed system. The high classification accuracies, especially for texture discrimination, imply that the proposed system may be capable of recognizing more complex textures and degrees of stiffness. Running tactile sensing experiments under more textures and varied stiffness may become an extension of this study as well.

REFERENCES

[1] R. Deimel and O. Brock, "A compliant hand based on a novel pneumatic actuator," *2013 IEEE Int. Conf. Robot. Autom.*, pp. 2047–2053, 2013.

[2] M. Giannaccini *et al.*, "A variable compliance, soft gripper," *Auton. Robots*, vol. 36, pp. 93–107, 2014.

[3] M. Tavakoli and A. T. Almeida, "Adaptive under-actuated anthropomorphic hand: ISR-SoftHand," *2014 IEEERSJ Int. Conf. Intell. Robots Syst.*, pp. 1629–1634, 2014.

[4] R. Deimel and O. Brock, "A novel type of compliant and underactuated robotic hand for dexterous grasping," *Int. J. Robot. Res.*, vol. 35, pp. 161–185, 2016.

[5] R. S. Johansson and J. R. Flanagan, "Coding and use of tactile signals from the fingertips in object manipulation tasks," *Nat. Rev. Neurosci.*, vol. 10, no. 5, pp. 345–359, May 2009, doi: 10.1038/nrn2621.

[6] R. Andrecioli and E. Engeberg, "Grasped object stiffness detection for adaptive PID sliding mode position control of a prosthetic hand," *2012*

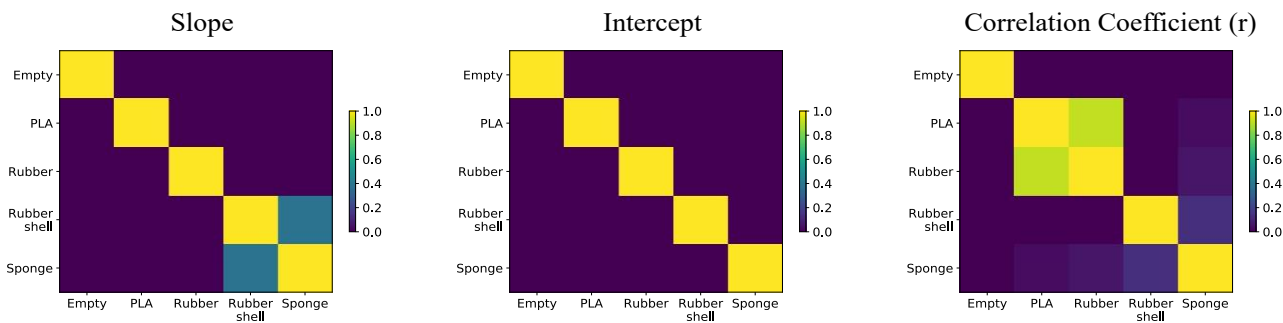

Fig. 10. The inverse symmetrical p-value matrices for stiffness features. The consistently low p-values between classes under these features indicate that the selected features can adequately represent difference between classes.

*4th IEEE RAS EMBS Int. Conf. Biomed. Robot. Biomechatronics BioRob*, pp. 526–531, 2012.

[7] H. Deng, X. Xu, W. Zhuo, and Y. Zhang, "Current-Sensor-Based Contact Stiffness Detection for Prosthetic Hands," *IEEE Access*, vol. 8, pp. 1–1, Feb. 2020, doi: 10.1109/ACCESS.2020.2972588.

[8] N. Jamali and C. Sammut, "Majority Voting: Material Classification by Tactile Sensing Using Surface Texture," *IEEE Trans. Robot.*, vol. 27, no. 3, pp. 508–521, Jun. 2011, doi: 10.1109/TRO.2011.2127110.

[9] K. Johnson, "The roles and functions of cutaneous mechanoreceptors," *Curr. Opin. Neurobiol.*, vol. 11, no. 4, pp. 455–461, Aug. 2001, doi: 10.1016/S0959-4388(00)00234-8.

[10] U. Rongala, A. Mazzoni, and C. Oddo, "Neuromorphic Artificial Touch for Categorization of Naturalistic Textures," *IEEE Trans. Neural Netw. Learn. Syst.*, Sep. 2015, doi: 10.1109/TNNLS.2015.2472477.

[11] H. Nguyen *et al.*, "Dynamic Texture Decoding Using a Neuromorphic Multilayer Tactile Sensor," in *2018 IEEE Biomedical Circuits and Systems Conference (BioCAS)*, Cleveland, OH, Oct. 2018, pp. 1–4. doi: 10.1109/BIOCAS.2018.8584826.

[12] S. Sankar *et al.*, "Texture Discrimination with a Soft Biomimetic Finger Using a Flexible Neuromorphic Tactile Sensor Array That Provides Sensory Feedback," *Soft Robot.*, p. soro.2020.0016, Sep. 2020, doi: 10.1089/soro.2020.0016.

[13] E. M. Izhikevich, "Simple model of spiking neurons," *IEEE Trans. Neural Netw.*, vol. 14, no. 6, pp. 1569–1572, Nov. 2003, doi: 10.1109/TNN.2003.820440.

[14] K. Hosoda, Y. Tada, and M. Asada, "Anthropomorphic robotic soft fingertip with randomly distributed receptors," *Robot. Auton Syst*, vol. 54, pp. 104–109, 2006.

[15] H. Zhao, K. O'Brien, S. Li, and R. F. Shepherd, "Optoelectronically innervated soft prosthetic hand via stretchable optical waveguides," *Sci. Robot.*, vol. 1, no. 1, p. eaai7529, Dec. 2016, doi: 10.1126/scirobotics.aai7529.

[16] J. C. Tuthill and E. Azim, "Proprioception," *Curr. Biol.*, vol. 28, no. 5, pp. R194–R203, Mar. 2018, doi: 10.1016/j.cub.2018.01.064.

[17] [17] B. Homberg, R. K. Katzschmann, M. Dogar, and D. Rus, "Haptic identification of objects using a modular soft robotic gripper," *2015 IEEERSJ Int. Conf. Intell. Robots Syst. IROS*, pp. 1698–1705, 2015.

[18] Y. Yan, Y. Wang, X. Chen, C. Shi, J. Yu, and C. Cheng, "A tendon-driven prosthetic hand using continuum structure*," in *2020 42nd Annual International Conference of the IEEE Engineering in Medicine Biology Society (EMBC)*, 2020, pp. 4951–4954. doi: 10.1109/EMBC44109.2020.9175357.

[19] M. Kutzer, S. Segreti, C. Brown, R. Taylor, and S. Mears, *Design of a new cable-driven manipulator with a large open lumen: Preliminary applications in the minimally-invasive removal of osteolysis*. 2011, p. 2920. doi: 10.1109/ICRA.2011.5980285.

[20] L. Qin and Y. Zhang, "Roughness discrimination with bio-inspired tactile sensor manually sliding on polished surfaces," *Sens. Actuators Phys.*, vol. 279, pp. 433–441, Aug. 2018, doi: 10.1016/j.sna.2018.06.049.

[21] A. Bicchi, "Hands for dexterous manipulation and robust grasping: a difficult road toward simplicity," *IEEE Trans. Robot. Autom.*, vol. 16, no. 6, pp. 652–662, Dec. 2000, doi: 10.1109/70.897777.